# A MULTI-STREAM HMM APPROACH TO OFFLINE HANDWRITTEN ARABIC WORD RECOGNITION

Ahlam MAQQOR, Akram HALLI, and Khaled SATORI

Laboratory LIIAN, University Sidi Mohamed Ben Abdellah Faculty of Science Dhar EL Mahraz, Fez, Morocco

ahlamhfl@gmail.com akram_halli@yahoo.fr khalidsatorim3i@yahoo.fr

### *ABSTRACT*

*In This paper we presented new approach for cursive Arabic text recognition system. The objective is to propose methodology analytical offline recognition of handwritten Arabic for rapid implementation.*
*The first part in the writing recognition system is the preprocessing phase is the preprocessing phase to prepare the data was introduces and extracts a set of simple statistical features by two methods : from a window which is sliding long that text line the right to left and the approach VH2D (consists in projecting every character on the abscissa, on the ordinate and the diagonals 45° and 135°) . It then injects the resulting feature vectors to Hidden Markov Model (HMM) and combined the two HMM by multi-stream approach.*

## 1. INTRODUCTION

Writing arable is naturally cursive, their recognition is the dream of everyone who needed data entry in a computer. Several solutions have been proposed for this recognition, quoting among others [9] [10] [11] [12] [13], these systems based MMCs have been developed for the recognition of cursive Arabic, there are several fields of applications in which the recognition eagerly awaited, such as automatic sorting of mail, automatic processing of administrative records, forms, surveys, or the registration of bank checks and postal.

The Arabic writing poses many problems for systems of automatic recognition we retain essentially these problems: the segmentation of handwritten words, skew angle of lines, overlaps, ligatures, the spaces between words, the dots are positioned above or below the character body and can change the meaning of the word.

To overcome these problems, an analytical approach has been proposed, The main objective of our system is to design and implement a multi-stream approach of two types of feature extraction based on local densities and configurations of pixels and features a projection based on vertical, horizontal and diagonal 45 °, 135 ° (VH2D approach), these characteristics is considered independent of the others and the combination is in a space of probability (combine the outputs of classifiers with a creation of a system of higher reliability ), in [15] [16] showed the combination of classifiers for the recognition of handwriting.

The main advantages of multi-stream models [17]:

- They provide a means to combine different sources of information.
- The topology of the hidden Markov Models may be adapted to each source of observation.
- Different flows can become out of sync.

In this paper, we discuss the characteristics of a handwritten Arabic script. Then, the image preprocessing to recognize the word, the segmentation of text, then the feature extraction with use the technique of sliding windows and VH2D approach. Finally we describe the modeling MMC with a combination multi-stream.

## 2. CHARACTERISTIC OF ARABIC SCRIPT

Arabic script is different compared to other types of writing Latin, Chinese....By their own structure and binding mode to form a word.

The difficulties related to the morphology of writing:

- Arabic script is written from right to left in a cursive way.
- The Arabic alphabet consists of 28 characters.
- Arabic script is inherently cursive.
- Some characters in a word can be overlapped vertically and multiple characters can be combined vertically to form a ligature (Fig.1).
- Some Arabic characters use diacritics (a diacritic may be placed above or below the body of the character and change the meaning of the word). (Fig. 2).
- An Arabic word usually consists of one or more connected components (sub words) each containing one or more characters (Fig. 3).
- Each character can take four different forms (beginning, middle, end, isolated) depending on its position in the word (Fig. 4).
- Change in calligraphic styles, six different graphic styles (Fig. 5).

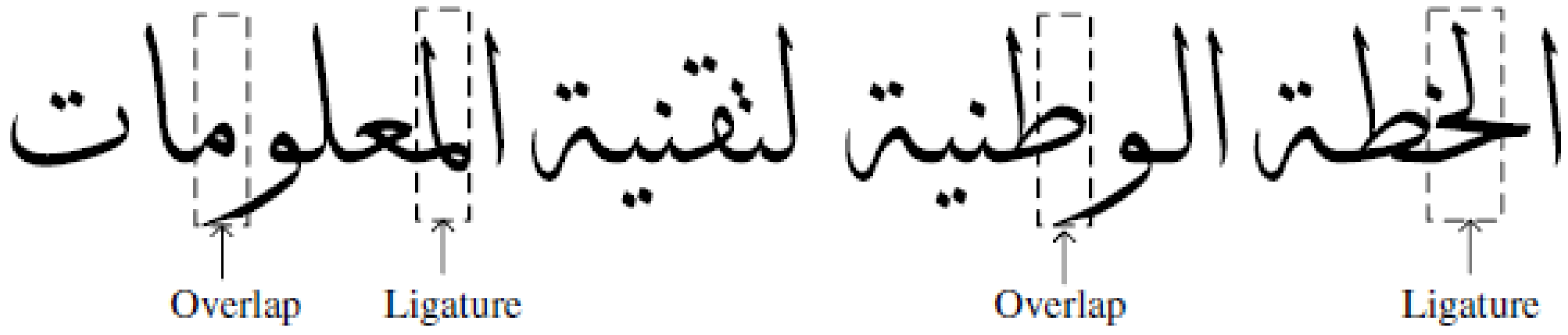

Fig.1 A sample of the “ligatures” and “overlaps”

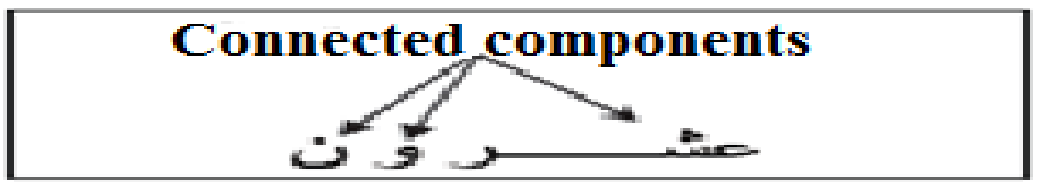

Fig.2 characters depending on the diacritical body

ب ث ت ن ر ز
ح ج خ

Fig.3 : Example of Arabic Word with connected components

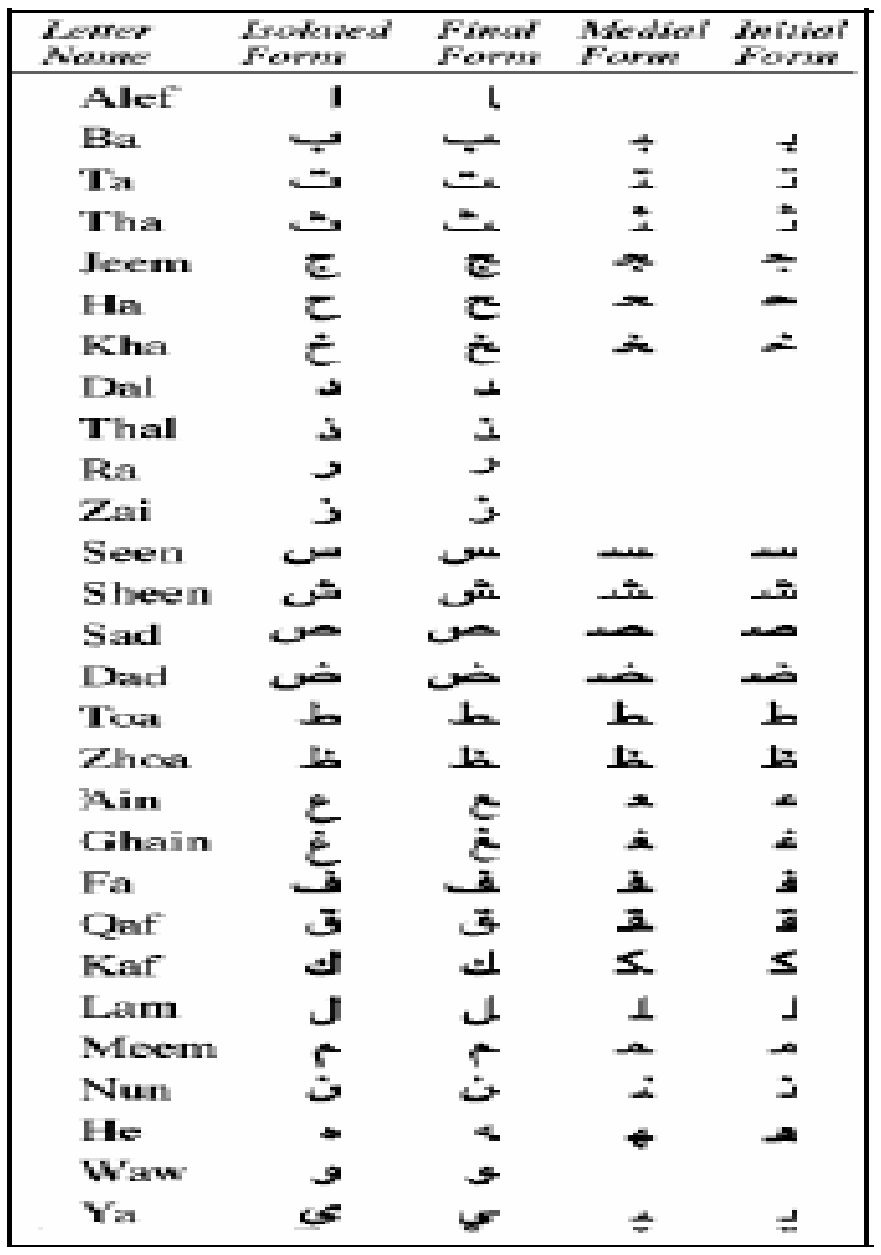

| Letter Name | Isolated Form | Final Form | Medial Form | Initial Form |
|---|---|---|---|---|
| Alef | ا | ـا | | |
| Ba | ب | ـب | ـبـ | بـ |
| Ta | ت | ـت | ـتـ | تـ |
| Tha | ث | ـث | ـثـ | ثـ |
| Jeem | ج | ـج | ـجـ | جـ |
| Ha | ح | ـح | ـحـ | حـ |
| Kha | خ | ـخ | ـخـ | خـ |
| Dal | د | ـد | | |
| Thal | ذ | ـذ | | |
| Ra | ر | ـر | | |
| Zai | ز | ـز | | |
| Seen | س | ـس | ـسـ | سـ |
| Sheen | ش | ـش | ـشـ | شـ |
| Sad | ص | ـص | ـصـ | صـ |
| Dad | ض | ـض | ـضـ | ضـ |
| Toa | ط | ـط | ـطـ | طـ |
| Zhoa | ظ | ـظ | ـظـ | ظـ |
| Ain | ع | ـع | ـعـ | عـ |
| Ghain | غ | ـغ | ـغـ | غـ |
| Fa | ف | ـف | ـفـ | فـ |
| Qaf | ق | ـق | ـقـ | قـ |
| Kaf | ك | ـك | ـكـ | كـ |
| Lam | ل | ـل | ـلـ | لـ |
| Meem | م | ـم | ـمـ | مـ |
| Nun | ن | ـن | ـنـ | نـ |
| He | ه | ـه | ـهـ | هـ |
| Waw | و | ـو | | |
| Ya | ي | ـي | ـيـ | يـ |

Fig.4 Different forms of the characters according to their position in the word

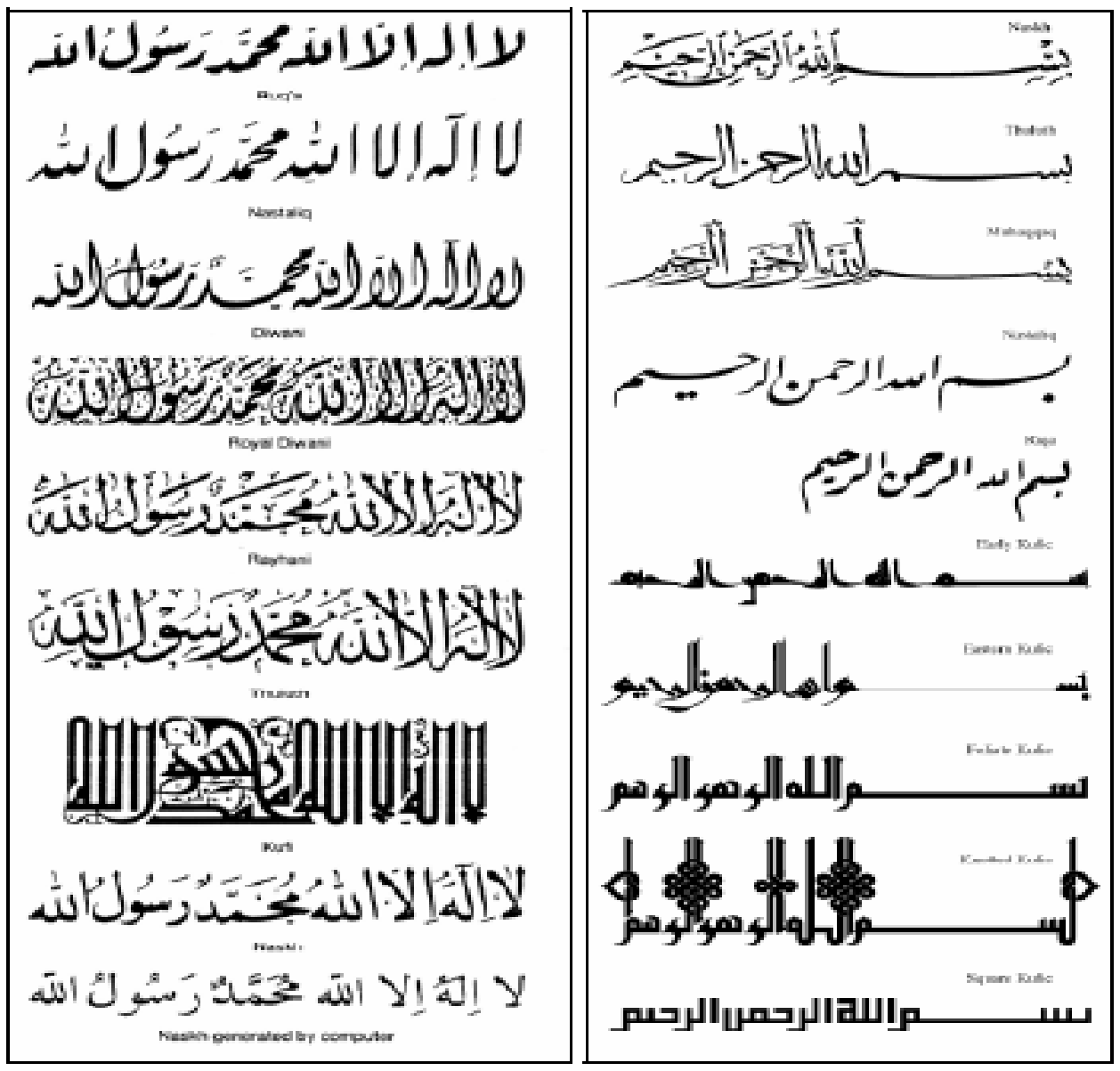

Fig.5 Variation of calligraphic style

## 3. RECOGNITION SYSTEM

The proposed recognition system is based on a model multi-stream (Fig.6). The system input is a word. The first step is to apply a series of preprocessing operations attempt to enhance the text image, to achieve this objective, a set of operation including thresholding, filtering, sommthing and skew detection. The next step be segmentation of the text image into lines, we focus on the separation of words for each line. Then the most important step of the recognition is feature extraction. Each word is shown in two sets of characteristics with the use of two different methods (Sliding Window approach and VH2D approach). Both sets characteristics extraction used by HMM recombination, for recognition the words a standard Viterbi Algorithm is used.

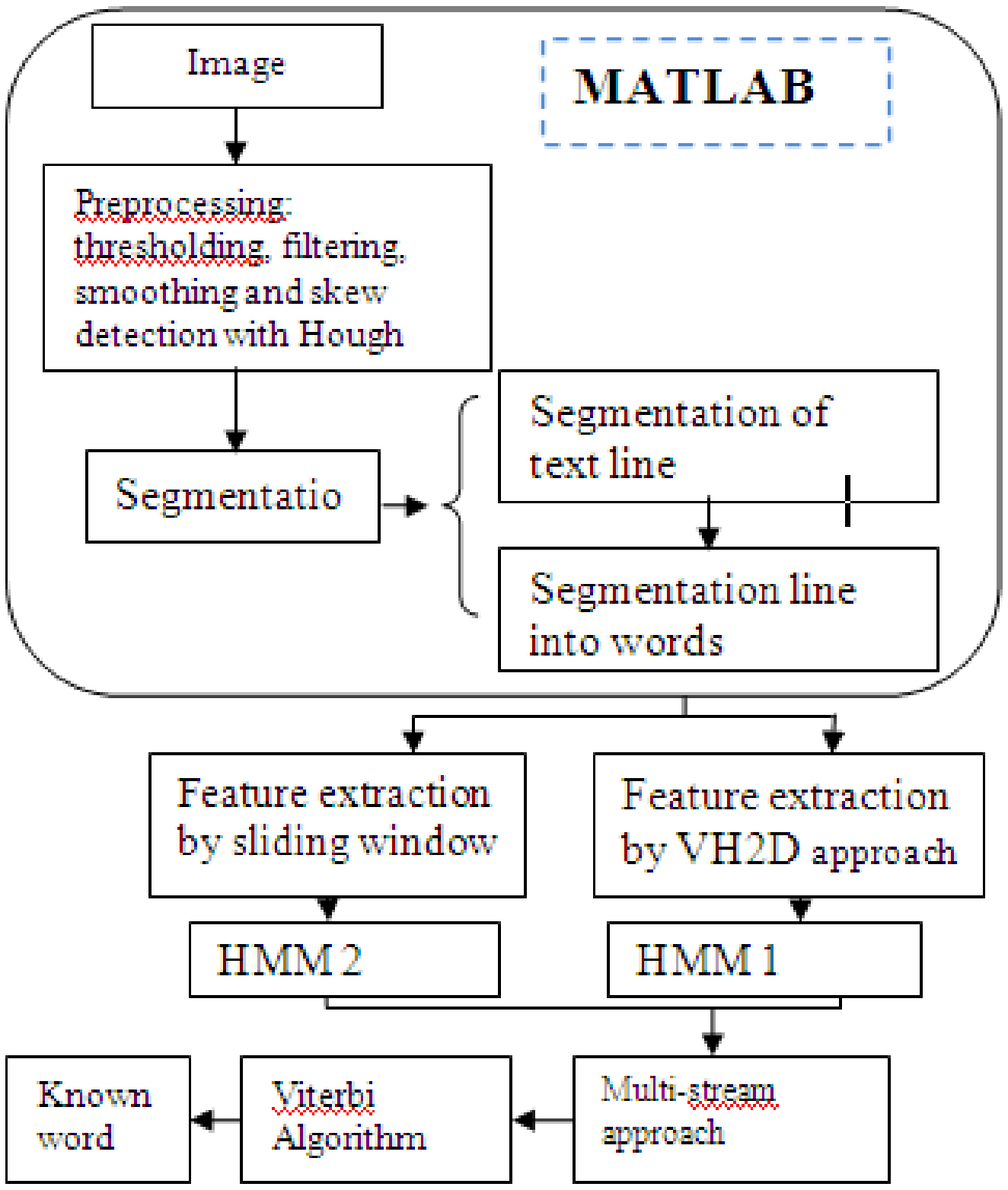

Fig.6 Description of the recognition system

## 4. PRE-PROCESSING

The text is scanned and stored as a binary image. The preprocessing applied to the image of the word can, firstly, to eliminate or reduce noise in the image and detects the skewing angle for align the text image in true horizontal.

To reach this purpose, in this paper a set of operations are applied to the text image including thresholding or binarization, filtering, smoothing, normalization and skew detection are applied to text image for simplify the feature extraction.

In our approach the median filter is applied to reduce the image noise. Simply, median filter can be of any central symmetric shape, a round disc, a square, a rectangle, or a cross. The pixel at the center will be replaced by the median of all pixel values inside the window. See Figure 7

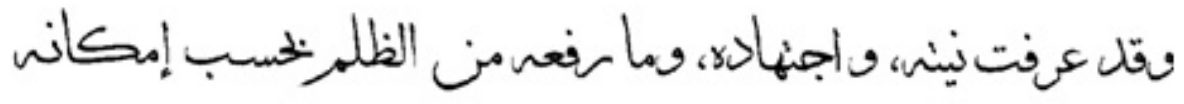

**a**

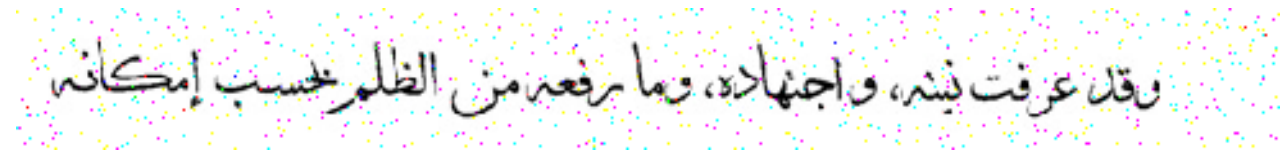

**b**

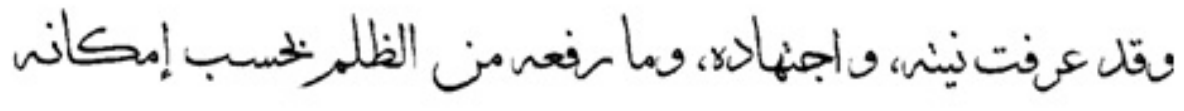

**c**

Fig.7 : a) original image without noise. b) image after adding a desired noise of type “salt & pepper”. c) image after using median filter

Once the text page is scanned, the text lines are inclined to the true horizontal axis. The most effective approach to estimate the angle of inclination is to use the Hough transform [14]. Hough transform is a method of detecting fragmented lines in a binary image.

Once the skew angle is determined, the text image is then rotated with the skewing angle in the reverse direction, as show in Figure

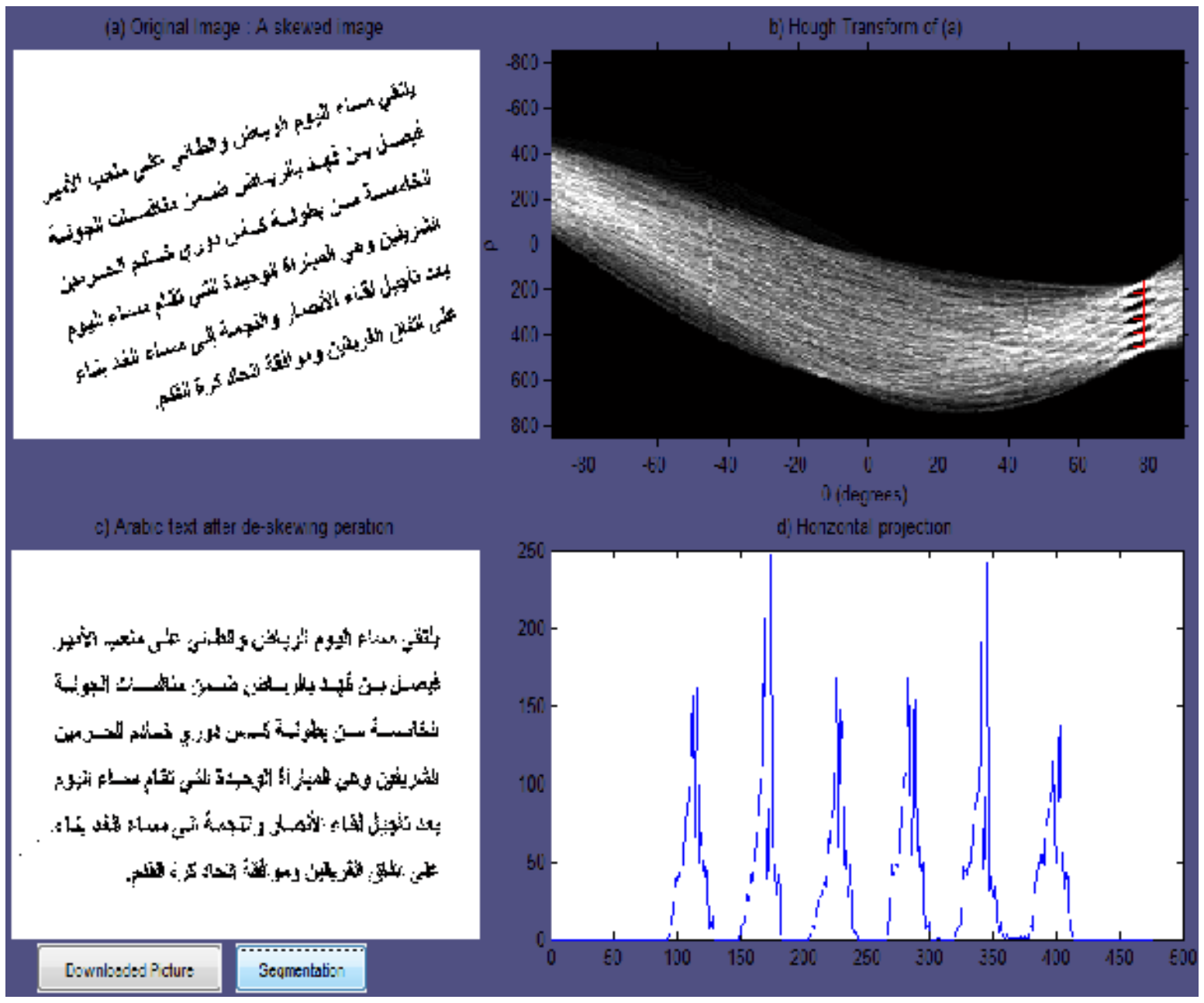

Fig.8 : a) the text in the image is skewed by 18° from the true horizontal axis. b) the Hough Transform of image in a) detects the skewing angle. c) Arabic text after de-skewing operation. d) Horizontal projection of (c).

Each text image in the data corpus is transferred into a set of line images with horizontal projection methods as described above, in our approach the extraction of the base line of writing the word is based on the method described in [1] It is based on the analysis of the histogram of horizontal projection

يلتقي مساء اليوم الرياض والطائي على ملعب الأمير

Line 1

فيصل بن فهد بالرياض ضمن منافسات الجولة

Line 2

Fig.9 : horizontal projection methods are used to find the line image

## 5. Segmentation

The most important step of automatic segmentation of documents is the text image lines in the future studies conducted based on a decomposition of the image connected components [1] [2] [3].

We used in our system a segmentation of the text image in line with use of the method of the horizontal projection that is nothing more than a simple sum of the number of pixels on each line. We can say the beginning or end of a text line is detected, if the value of horizontal projection is below a threshold (the threshold is picks in the histogram).

After separation of the lines, we focus on the separation of words for each line

## 6. Feature extraction

We used two methods in our system for the extraction of features:

### 6.1. Sliding Window

Each picture of the word is transformed into a sequence of feature vectors extracted from right to left on binary images of words by sliding window of size N pixel successively offset and £ of pixels (£ parameter that takes values between 1 and N-1). Each window is divided vertically into a number of fixed cells.

These extracted features are divided into two families: the characteristics of local densities and configuration of the pixels.

The advantage of using this type of feature are independent of the language used and also can be used for any type of cursive including poles and legs like writing Arabic and Latin writing.
The characteristics of the densities of pixels:

- F1: density of black pixels in the window.
- F2: density of white pixels in the window.
- F3: Number of black / white transitions between cells.
- F4: difference between position the center of gravity G the pixels for two consecutive Sliding Windows.
- F5 to F12 are the densities of pixels in each writing column of the window.
- F13: center the gravity the pixels writing.

The characteristics of the local configurations of pixels:

- F14 to F18: The number of white pixels that belong to one of five configurations of the figure in each window.

The result of this step is a feature vector with 18 features.

### 2.6. VH2D approach

The VH2D approach proposed in (Xia and Cheng, 1996) consists in projecting every character on the abscissa, on the ordinate and the diagonals 45° and 135°.

The projections take place while calculating the sum of the values of the pixels $i_{xy}$ according to a given direction

#### 2.6.1. Presentation of the VH2D

a) *Vertical projection* : The vertical projection of an image I=Ixy (of dimension N x N) representing a character C is indicated by :

$$P^{v} = [P_1^{v}, P_2^{v}, p_3^{v} \dots, P_{N-1}^{v}, P_N^{v}] \; where \; P_y^{v} = \sum_{x=1}^{N} i_{xy}$$

b) *Horizontal projection:* the projection of an image I=Ixy (of dimension N x N) representing a character C is indicated by:

$$P^{h} = [P_1^{h}, P_2^{h}, \dots, P_{N-1}^{h}, P_N^{h}] \; where \; P_y^{h} = \sum_{y=1}^{N} i_{xy}$$

c) *Diagonal projection* (45°): the projection of an image I=Ixy (of dimension N x N) representing a character C is indicated by:

$$P^{d1} = \left[P_1^{d1}, P_2^{d1}, p_3^{d1}, \dots, P_{N-1}^{d1}, P_N^{d1}\right] \quad where :$$

$$P_m^{d1} = \begin{cases} \sum_{l=N-m+1}^{N} \sum_{k=1}^{m} i_{lk} & 1 \le m \le N \quad et\; l = k + N - m \\ \sum_{l=1}^{2N-m} \sum_{k=m-N+1}^{N} i_{lk} & N+1 \le m \le 2N-1 \; et\; l = k + N - m \end{cases}$$

*Projection on the diagonal 135°* : Projection on the diagonal 135° of an image I=Ixy (of dimension N x N) representing a character C is indicated by :

$$P_m^{d2} = \begin{cases} \sum_{l=1}^{m} \sum_{k=1}^{m} i_{lk} & 1 \le m \le N \quad et\; k = m - l + 1 \\ \sum_{l=m-N+1}^{2N-m} \sum_{k=m-N+1}^{N} i_{lk} & N+1 \le m \le 2N-1 \quad et\; k = m - l + 1 \end{cases}$$

## 7. MULTI-STREAM APPROACH

The multi-stream initially proposed in [7] [8] is an adaptive method for combining different sources of information using Markov models.

The multi-flow method is intended to fill these gaps. It includes three types of fusion :

### 7.1 Multi-classifiers

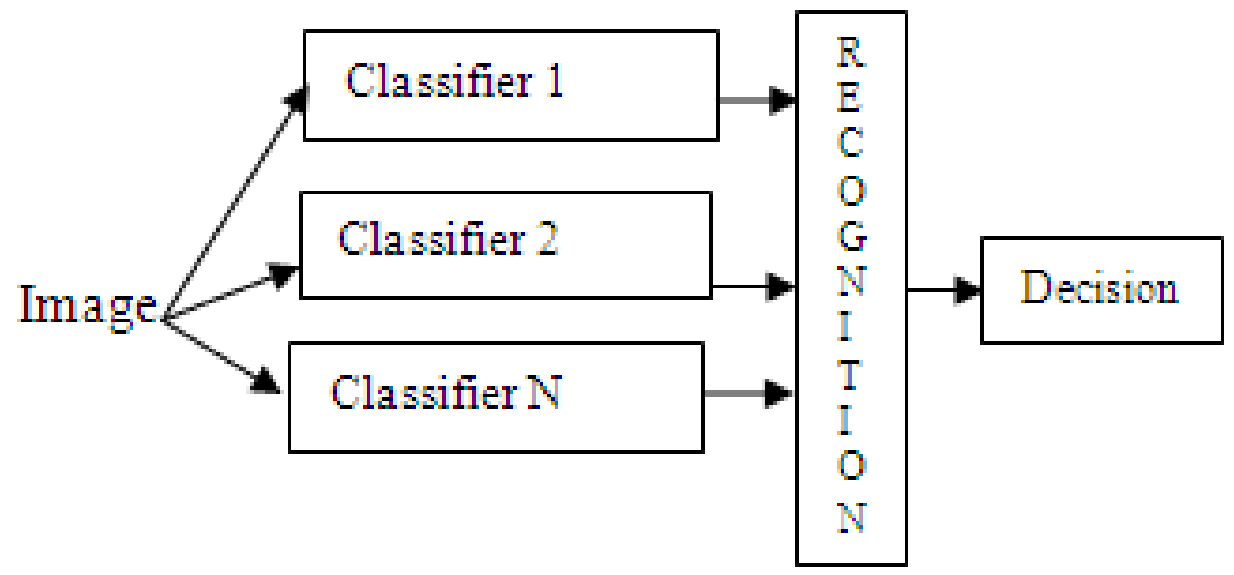

### 7.2 Multi-modal approach

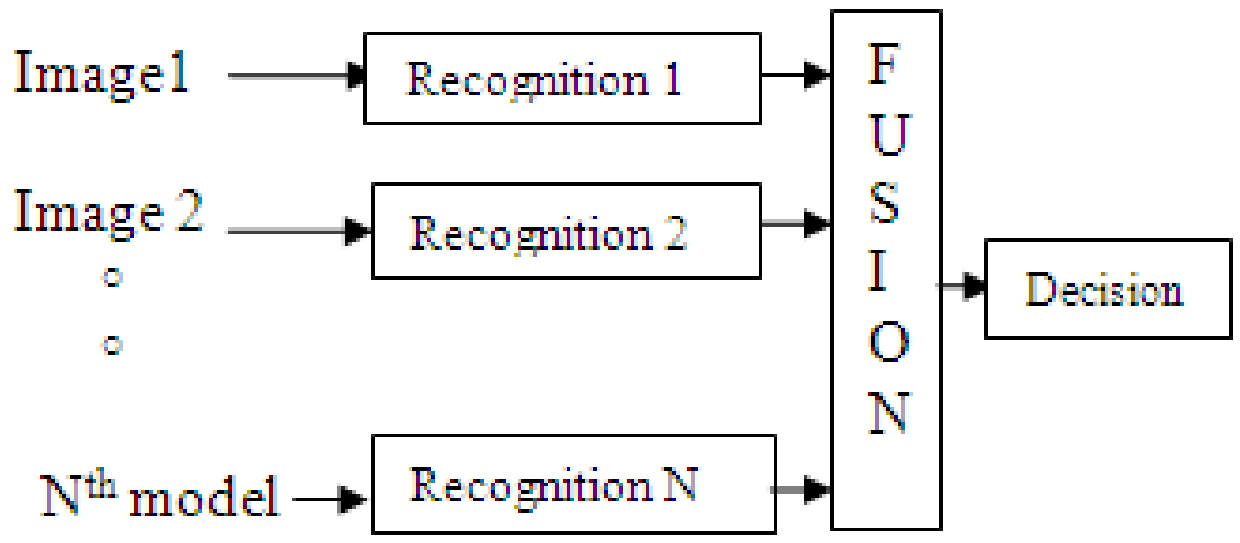

### 7.3 Partial recognition approach or multi-band

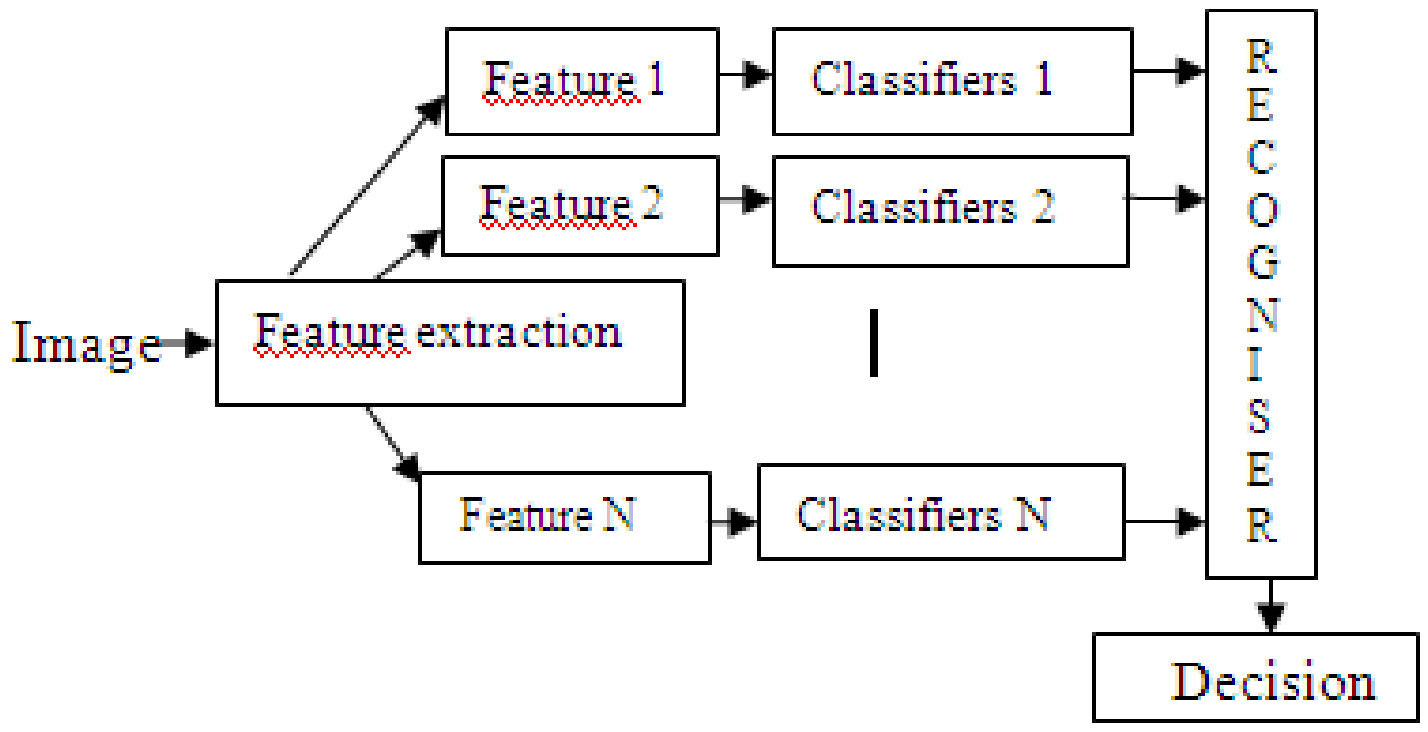

Fig.10: The method consists of three multi-stream fusions: Multi-classifier approach, the multi-modal approach and partial recognition or multi-band

In our case we chose the partial recognition approach or multi-band that there are two types of data from the text image of the word.

## 7.4 Multi-stream formalism

K is the number of information sources and M is the model consists sequence of J (eg letters) in models that correspond to lexical items in $M_j$ (j = 1,2, ... ... N). Each sub model M consists of K Markov models (HMM) independent $M_j^k$(k=1,2,…A)

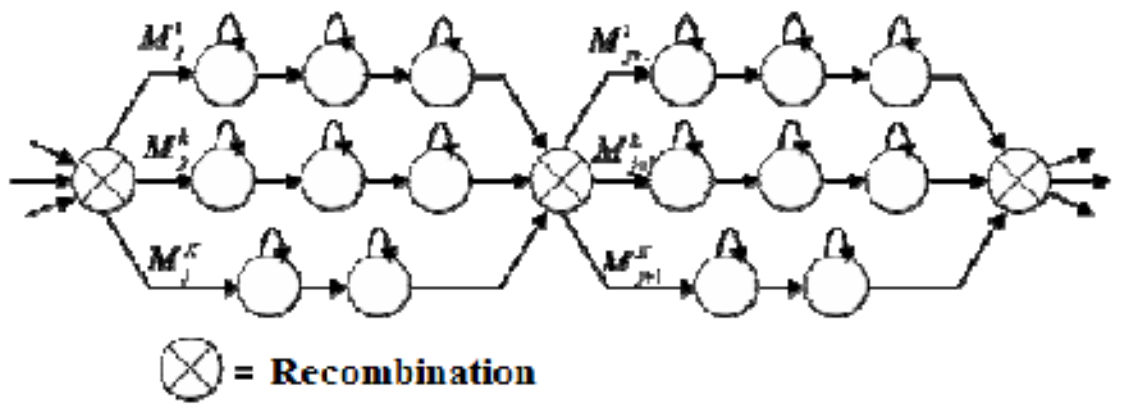

Fig. 11: Multi-Stream model

### 7.4.1. Definition of the multi-stream Model

$C_j$ is sequence of states associated with multi-stream sequence of multi-stream observations Xj, the probability P ($X_j$, $C_j$ | $M_j$) is calculated on the basis of the likelihoods associated with each of the sources of information according to the following :

$$P(X_j, C_j | M_j) = f(\{P(X_j^k, C_j^k | M_j^k), k = 1,2, \dots \dots \dots k\})$$

$f$ is function of linear combination

$X_j^k$ is sequence of observation vectors associated with the flow k

$C_j^k$ is sub-sequence of states associated with the flow k pattern in the $M_j^k$

The likelihood of the subsequence from the model Xj subunit Mj and Cj the path is written

$$\log P(X|M) = \sum_{j=1}^{N} \sum_{k=1}^{A} W_j^k \log P(X_j^k | M_j^k)$$

$W_j^k$ Represents the reliability of the flow k.

The models multi-stream amounts to calculating the likelihood of the best path in the sequence of vectors of observations of different sources of information with the Viterbi Algorithm.

## 8. HMM-RECOMBINATION

In opposition to printed text in most languages, the characters in cursive handwritten words are connected because of it several methods use Hidden Markov Models (HMM) for recognising handwritten words, have been very successfully.

system models words and characters in the form of Hidden Markov Models  is analytical: the models words are built by concatenation of models the characters. that are left-right as shown in Fig. 12 [4][5][6] gives an example of the training, showing that each character shape of the same type, independent of the word where it was written, contributes to the statistical character shape model. This enables a statistical training with less training data than in the case of word based models.

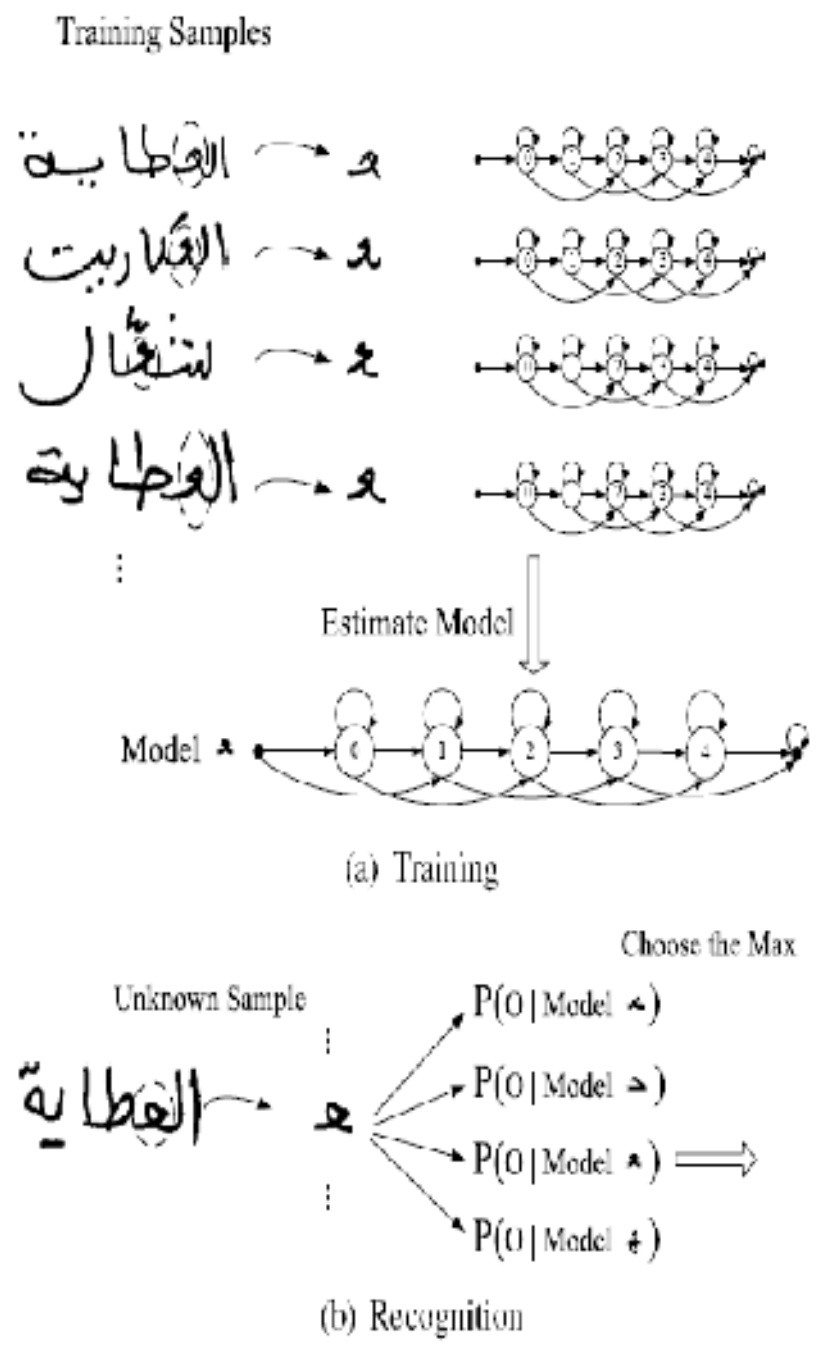

Fig.12 : HMM is trained for each “mode” using a number of examples of that “mode” from the training set, and (b) to recognize some unknown sequence of “modes”, the likelihood of each model generating that sequence is calculated and the most likely (maximum) model identifies the sequence.

"HMM-recombination" [18]: This algorithm is an adaptation of the algorithm "HMM-decomposition" [19] used in the field of speech recognition for decomposing the audio signal into two independent components (sound and noise). The idea is to build a composite MMC from MMC parallel multi-stream(Fig 10).

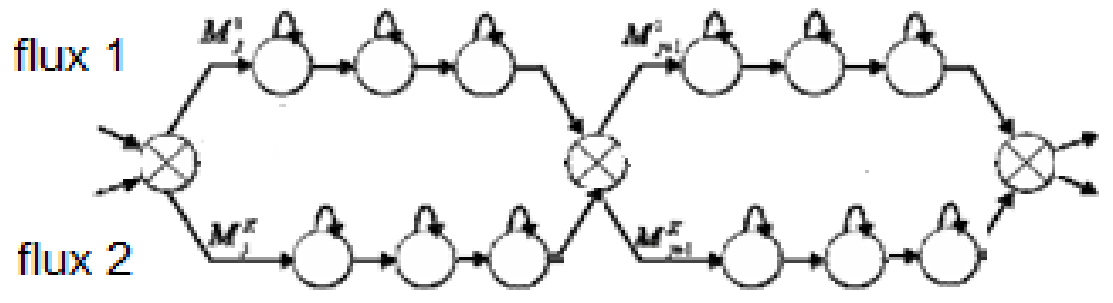

Fig.13: Parallel Model for 2 Stream

## 9. Experiences and Results

Evaluation of the performance of the proposed method is performed on a 200 word.we present the results obtained for the recognition of models with two type of feature extraction sliding window method and VH2D approach.

Table 1: Recognition results

| Recognizer | Recognition rate |
|---|---|
| Sliding Window (1) | 78.2 |
| VH2D (2) | 76.6 |
| Multi-flux(combination of classifiers 1 and 2) | 83.8 |

Recognition results of multi-stream model are the best, while the recognition rate is not so high considering the poor quality of writing certain words based.

## 10. CONCLUSION

We have presented a recognition system off-line handwritten Arabic script based on a multi-stream with Hidden Markov Models. The multi-stream models proposed method is illustrated the advantage of extracting primitive vectors by vertical windows and approach VH2D, Our system considers that the VH2D is sufficient because the recognition will be confirmed with two classifiers.

Mr. Khalid Satori received the PhD degree from the National Institute for the Applied Sciences INSA at Lyon in 1993. He is currently a professor of computer science at USMBA-Fez University. His is the director of the LIIAN Laboratory. His research interests include real-time rendering, Image-based rendering, virtual reality, biomedical signal, camera self calibration and 3D reconstruction.

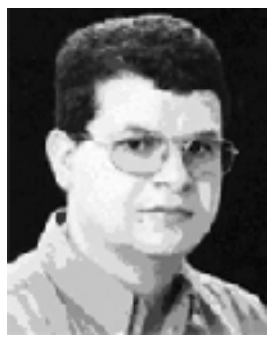

Mr. Akram Halli received the bachelor's and master's degrees from USMBA-Fez University in 2002 and 2004 respectively. He is currently working toward the PhD degree in the LIIAN Laboratory at USMBA-Fez University. His current research interests include real-time rendering, Image-based rendering and virtual reality.

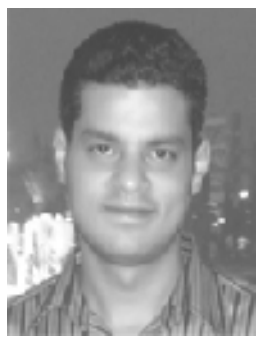

Miss. AHLAM MAQQOR received the bachelor's and mitrise's degrees from FST-Fez University in 2001 and 2007 respectively. She is a Ph.D student from Sidi Mohamed Ben Abdellah University (Fez-Morocco), her research interests include Handwritten Arabic Word recognition, image processing, Hidden Markov Models, etc.

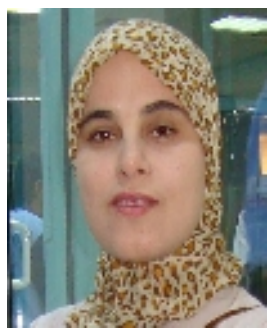